\documentclass[10pt,twocolumn,letterpaper]{article}

\usepackage{cvpr}              % To produce the CAMERA-READY version
\usepackage{wrapfig}

\usepackage{xspace}
\makeatletter
\DeclareRobustCommand\onedot{\futurelet\@let@token\@onedot}
\def\@onedot{\ifx\@let@token.\else.\null\fi\xspace}

\makeatother

\newcommand{\bB}{\mathbf{B}}
\newcommand{\bC}{\mathbf{C}}

\newcommand{\bM}{\mathbf{M}}

\newcommand{\ba}{\mathbf{a}}
\newcommand{\bb}{\mathbf{b}}
\newcommand{\bc}{\mathbf{c}}

\newcommand{\be}{\mathbf{e}}
\newcommand{\bff}{\mathbf{f}}

\newcommand{\bp}{\mathbf{p}}
\newcommand{\br}{\mathbf{r}}
\newcommand{\bs}{\mathbf{s}}

\newcommand{\bv}{\mathbf{v}}
\newcommand{\bx}{\mathbf{x}}

\newcommand{\bq}{\mathbf{q}}
\newcommand{\bw}{\mathbf{w}}

\usepackage[dvipsnames]{xcolor}

\definecolor{cvprblue}{rgb}{0.21,0.49,0.74}
\usepackage[pagebackref,breaklinks,colorlinks,citecolor=cvprblue]{hyperref}

\def\paperID{240} % *** Enter the Paper ID here
\def\confName{3DV\xspace}
\def\confYear{2026\xspace}

\title{Momentum-Conserving Graph Neural Networks for Deformable Objects}

\author{Jiahong Wang$^{1}$ \quad
Logan Numerow$^{2}$ \quad 
Stelian Coros$^{2}$ \quad
Christian Theobalt$^{1}$ \\
Vahid Babaei$^{1,3,4}$ \quad
Bernhard Thomaszewski$^{2}$\\
$^{1}$Max Planck Institute for Informatics\quad
$^{2}$ETH Z\"{u}rich\quad
$^{3}$University of Bonn\\
$^{4}$Fraunhofer Institute for Algorithms and Scientific Computing
}

\begin{document}
\maketitle

\begin{abstract}
Graph neural networks (GNNs) have emerged as a versatile and efficient option for modeling the dynamic behavior of deformable materials. While GNNs generalize readily to arbitrary shapes, mesh topologies, and material parameters, existing architectures struggle to correctly predict the temporal evolution of key physical quantities such as linear and angular momentum. In this work, we propose MomentumGNN---a novel architecture designed to accurately track momentum by construction. Unlike existing GNNs that output unconstrained nodal accelerations, our model predicts per-edge stretching and bending impulses which guarantee the preservation of linear and angular momentum. We train our network in an unsupervised fashion using a physics-based loss, and we show that our method outperforms baselines in a number of common scenarios where momentum plays a pivotal role.
\end{abstract}
    
\section{Introduction}
Simulating the dynamics of deformable materials is a long-standing research problem in computer graphics and engineering \cite{10.1145/37401.37427,10.1145/566654.566581, 10.1145/2601097.2601116, 10.1145/2994258.2994272, 10.1145/3386569.3392425}. The applications range from video games and animated movies to soft robotics, surgery training, and packaging design. 
Although established simulation methods based on, e.g., finite elements are robust, versatile, and accurate, the deep learning breakthrough led to the rise of (neural) surrogate models with the goal of learning and replacing computationally expensive physics-based simulations. 
By learning the mapping between external forces or constraints and resulting deformations, these surrogate models can offer substantial speedups while maintaining physical plausibility.

Among the various neural architectures, graph neural networks (GNNs) stand out for their ability to model complex physical systems. 
\begin{figure}[t]
% \vspace{0.1cm}
\begin{minipage}{\columnwidth}
    \centering
    \includegraphics[width=0.85\columnwidth]{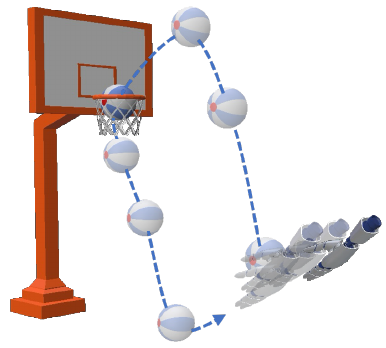}
    \caption{Simulation of a basket-shooting scene, where accurate modeling of momentum plays a crucial role. Our neural simulation method, \textit{MomentumGNN}, accurately reproduces the momentum evolution from a ground truth simulation, leading to a scored shot. (More details in Fig. \ref{fig:basket})}
    \label{fig:teaser}
\end{minipage}
\vspace{-0.8cm}
\end{figure}
A landmark work in this area is MeshGraphNets \cite{meshgraphnet}, which introduced the use of GNNs for deformable object simulation via an encode-process-decode architecture that takes a simulation mesh as input and predicts nodal accelerations. This framework has since inspired a range of follow-up studies demonstrating compelling results in cloth simulation \cite{Libao23MeshGraphNetpp}, garment modeling \cite{grigorev2022hood, Grigorev24ContourCraft}, and character animation \cite{Zheng_2021_CVPR}.
Despite their promise, existing graph-based approaches often struggle to conserve fundamental physical quantities, most notably momentum, because of their reliance on implicit, data-driven learning. They typically approximate physical dynamics by observation data without explicitly enforcing physical constraints. As a result, they exhibit non-physical behaviors such as drift and unnatural spin, which is particularly problematic in tasks involving free motion and collisions, where accurate momentum conservation is crucial for realism and reliability.

To address these limitations, we propose a novel GNN architecture that enforces physically-correct momentum evolution by design. Our main insight is to decompose momentum changes into contributions from external non-conservative forces, and those arising from internal elastic forces that conserve momentum. 
Our network then predicts corrective impulses that conserve momentum, effectively refining the state produced by the momentum step.
Specifically, we modify MeshGraphNets by replacing per-vertex decoders by \textit{per-edge} decoders that predict momentum-conserving bending and stretching impulses.
One key component is a novel layer-by-layer architecture that sequentially updates vertex positions using momentum-conserving impulses at each network layer, enabling greater representational capacity and improved physical fidelity. Briefly, our main contributions are: 
\begin{itemize}
    \item A physics-aware graph neural network that conserves the momentum by working on per-edge impulses. 
    \item A layer-by-layer architecture in which each network layer sequentially updates vertex positions through momentum-conserving impulses.
\end{itemize}

\section{Related Work}
\paragraph{Neural Simulation}
Unlike traditional machine learning problems where inputs and outputs, e.g., images, are uniform in size and regularly structured, deformable object simulations have variable numbers of vertices and are highly complex in connectivity. One potential solution is to infer node-wise corrections to pre-computed linear nodal displacements \cite{8536417}. Another strategy is model reduction \cite{Fulton:LSD:2018,10.1145/3450626.3459875}. where neural networks encode mesh to a reduced-size latent embedding, which speeds up numerical simulations. More recent improvements involve the use of higher-order derivatives \cite{high-order-diff} and self-supervised physics-based energy loss \cite{Wang_2024_CVPR, datafree}.

Neural methods have received particular attention in cloth simulation \cite{drapenet}. In the specific domain of garments, clothes are encoded as displacement offsets from the body \cite{cloth3d}. To combat over-smoothing artifacts, various methods are proposed, such as decomposition of low-frequency deformations into high-frequency details \cite{tailornet} and wrinkles with coarse meshes \cite{deepwrinkles}. Other recent methods \cite{deepsd, pbns} further exploit the parameterization of body objects and take advantage of the well-established SMPL parameterization and Pose Space Deformation (PSD). SNUG \cite{snug} and Neural Cloth Simulation (NCS) \cite{ncs} extend single-frame prediction to a continuous dynamics by embedding recurrent units into the network.

\paragraph{GNNs}
Despite impressive results, aforementioned models are limited in generalization ability. They are either subject to a fixed topology or rely on body parameterizations, which restrict them from broader applications in arbitrary shapes and dynamic topology changes. To resolve these drawbacks, mesh-based neural methods using graph neural networks (GNN) have become increasingly popular. 

Graph Networks (GN) \cite{pmlr-v80-sanchez-gonzalez18a} were initially proposed for rigid systems in skeleton robotics. They were further extended to an encode-process-decode architecture for general fluid particle systems \cite{pmlr-v119-sanchez-gonzalez20a}. Arguably, the most important work in mesh-based methods is MeshGraphNets \cite{meshgraphnet}, as it inaugurates mesh-based methods in deformable objects. It borrows the encode-process-decode ideas and upgrades GN blocks with residual connections. It has shown stunning results in modeling the cloth dynamics in the wind and simple collisions. Later works in this area are derived mostly from MeshGraphNets, where they add random edge connections \cite{gladstone2024mesh} or construct hierarchical U-net-like graph architectures \cite{grigorev2022hood, gladstone2024mesh} to facilitate long-distance propagation, followed by further improvements in multi-layer collision handling in garments \cite{Grigorev24ContourCraft}. MeshGraphNetRP \cite{Libao23MeshGraphNetpp} furthermore adds RNN-based state encoding to reproduce cloth oscillations. Deep Emulator \cite{Zheng_2021_CVPR} applies mesh-based simulation to improve cage-based deformations of animated characters, where they use GNNs to predict the secondary dynamics of cages surrounding the character. 
Compared with existing mesh-based methods, our work focuses on a different objective, i.e., momentum conservation.

\paragraph{Physics-Inspired Networks} 
Initial explorations into leveraging neural networks to accelerate physical simulation can be traced back to the 1990s \cite{grzeszczuk1998neuroanimator}. More recently, the emergence and widespread adoption of physics-informed neural networks (PINN) \cite{pinn} in solving partial differential equations (PDEs) have reignited significant interest in this domain. The resurgence has been particularly impactful in computational fluid dynamics (CFD), enabling advances such as surface pressure fields estimation \cite{HINES2023108268}, solutions to the Reynolds-Averaged Navier–Stokes equations \cite{doi:10.2514/1.J058291}, the acceleration of traditional turbulence solvers  \cite{10.1145/3392717.3392772}, and interactive aerodynamics design tools \cite{NEURIPS2022_59593615}. Concurrently, there is another line of research in approximating energy-like quantities such as Hamiltonian \cite{NEURIPS2019_26cd8eca} and Lagrangian \cite{lutter2018deep} by neural networks. They are further improved with graph neural networks \cite{sanchez2019hamiltonian} and decompositions of dissipative and conservative dynamics \cite{sosanya2022dissipative}.
Unlike prior methods that emphasize PDEs and energy conservation, we focus on momentum conservation.
Although recent studies consider momentum-conserving dynamics for fluids \cite{10.5555/3600270.3600770} and multi-body particles \cite{sharma2026physics}, these approaches do not readily extend to solid and cloth systems governed by elastic potentials. We propose specialized stretching and bending impulses to handle these settings.

\section{Background}

Our work builds on MeshGraphNets, a mesh-based graph neural network (GNN) architecture trained to predict the dynamics of deformable materials. It implements an Encode-Process-Decode design: starting from the current configuration (e.g., vertex positions and velocities), per-vertex input features are passed through an MLP encoder, a GNN processor, and an MLP decoder to produce per-vertex accelerations. These accelerations are then integrated in time to obtain the vertex positions at the end of the time step. We explain each component in the following paragraphs.

\paragraph{Input graph} The input graph $G_\mathrm{in}=(V, E)$ is constructed from the initial simulation mesh $M^0$. A graph node ${\bx_i}$ is created for every mesh vertex, and a bidirectional graph edge $\be =(\be_{ij},\be_{ji})$ is added for every mesh edge. Furthermore, each node is endowed with a vertex feature $\br_i$ and every graph edge is given two edges features, $\bs_{ij}$ and $\bs_{ji}$, corresponding to the two uni-directional edges, respectively.

\paragraph{Encoder} The encoder includes a node MLP and an edge MLP, which take the input graph and lift input features such as vertex velocity and edge length to a higher-dimensional latent space. 

\paragraph{Processor} The processor is composed of $n_L$ message passing layers of graph network blocks \cite{pmlr-v80-sanchez-gonzalez18a}. Each layer learns a separate set of parameters and is applied sequentially to propagate information in local neighborhoods. The message passing begins by computing messages ${\bf m}_{ij}$ for every edge ${\bf e}_{ij}$ by an edge MLP $f^\mathrm{edge}$,
\begin{align}
    {\bf m}_{ij}\leftarrow f^\mathrm{edge}({\bf s}_{ij},{\bf r}_i,{\bf r}_j) \label{eqn:message} \ ,
\end{align}
where $\br_i$ and $\br_j$ are the vertex features of the two nodes belonging to edge $\be_{ij}$.
With all messages computed in this way, node and edge features are then updated as,
\begin{align}
    {\bf s}_{ij}\leftarrow {\bf s}_{ij} + {\bf m}_{ij}\ ,\\
    {\bf r}_i\leftarrow {\bf r}_i + f^\mathrm{vertex}({\bf r}_i,\sum_{j\in \mathcal{N}_i} {\bf m}_{ij}) \ ,
\end{align}
where the vertex MLP $f^\mathrm{vertex}$ aggregates incoming edge messages with vertex features.
Each layer in the processor block corresponds to a single message passing step. 
It is worth noting that, for a given message passing step,  all vertex and edge features are processed with the same vertex and edge MLP, respectively. However, every layer has its own vertex and edge MLP.

\paragraph{Decoder} After a given number of message passing steps, the vertex features from the last layer are passed into a decoder MLP which returns per-vertex accelerations ${\bf a}=(\ba_1,\ldots,\ba_{|V|})^T$. These accelerations are then integrated in time to obtain updated positions,
\begin{align}
    {\bf x}^{i+1} = {\bf x}^{i} + \Delta t{\bf v}^{i} + \Delta t^2 {\bf a} \ ,\label{eqn:euler_int}
\end{align}
from which velocities are updated using finite differences,
\begin{align}
\label{eq:velFD}
    {\bf v}^{i+1} &= \frac{1}{\Delta t} ({\bf x}^{i+1} - {\bf x}^{i}) \ .
\end{align}

\paragraph{Training} The MeshGraphNets architecture outlined above can be trained in a supervised way using curated simulation data \cite{meshgraphnet}. As shown by \citet{grigorev2022hood}, MeshGraphNets can also be trained in a self-supervised way when using a loss function that directly encodes the governing physics. Using the optimization-based formulation of implicit Euler \cite{example2} as a basis, the per-time-step potential is defined as
\begin{equation}
\label{eq:IELoss}
   E_\mathrm{IE}(\bx) = \frac{1}{2}\Delta \bv^T\bM \Delta \bv + E_\mathrm{int}(\bx) + \bff_\mathrm{ext}^T\bx\ ,
\end{equation}
where $\Delta \bv = \frac{1}{\Delta t}(\bx-\bx_i)-\bv_i$, $\bM$ is the mass matrix, $E_\mathrm{int}$ is the internal elastic energy, and $\bff_\mathrm{ext}$ denotes external (i.e., non-conservative) forces. Each minimizer of this potential,
\begin{equation}
    \bx^{i+1}=\arg \min_\bx \  E_\mathrm{IE}(\bx) \ ,
\end{equation}
is a solution to the implicit Euler equations.
Total loss is defined as the sum of per-step losses over all samples in the corresponding training set.

\paragraph{Discussion}
Regardless of the training strategy, existing MeshGraphNets architectures output per-vertex accelerations. Since these accelerations have no \textit{a priori} restrictions, they can induce arbitrary momentum changes. Our experiments show that spurious momentum changes are a general problem with existing MeshGraphNets architectures, particularly noticeable for ballistic motions. 
We conjecture that the reason for this nonphysical behavior is the fact that learning the relationship between vertex accelerations and global orientation of the system would require a dense sampling of rotational transformations and thus lead to excessive amounts of data. 
We propose a method that avoids this problem by restricting accelerations to the momentum-conserving subset by construction. 

\begin{figure*}[ht]
\vspace{-0.5cm}
\begin{minipage}{\linewidth}
    \centering
    \includegraphics[width=0.9\linewidth]{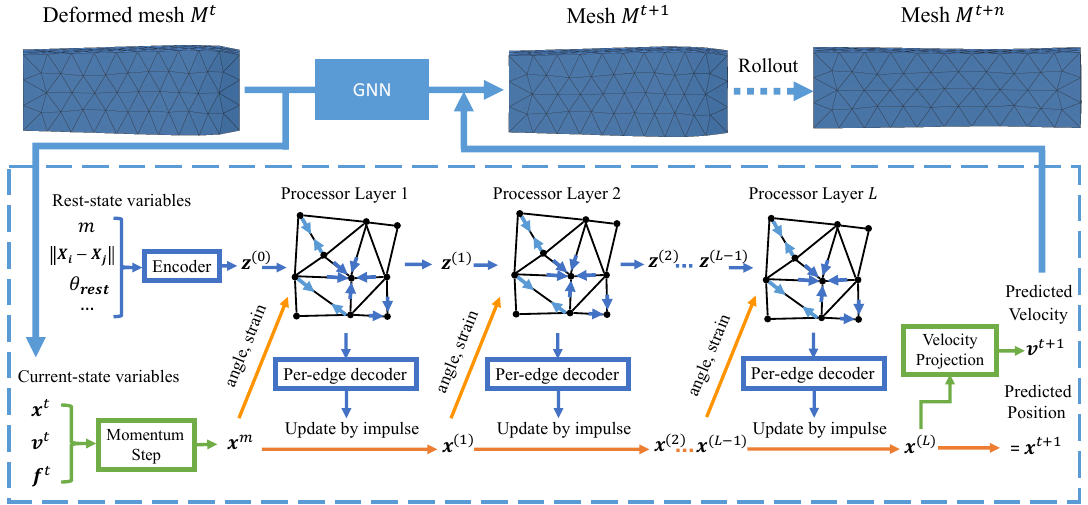}
    \caption{We visualize MomentumGNN and denote latent edge and node features jointly by ${\bf z}=\{{{\bf s}_{ij},{\bf r}_i}\}$. Starting from a {\it momentum step} that integrates contributions from inertia and non-conservative external forces, our network predicts momentum-conserving stretching and bending impulses in each layer and sequentially updates position ${\bf x}^{(l)}$ to obtain the final prediction.}
    \label{fig:overview}
\end{minipage}
\vspace{-0.6cm}
\end{figure*}

\section{Method}

The goal of our work is to develop a GNN architecture that accurately predicts the temporal evolution of linear and angular momentum. We start by separating discrete time steps into momentum-changing and momentum-conserving sub-steps (Sec. \ref{sec:MomentumStep}). We then introduce a basis for momentum-conserving impulses (Sec. \ref{sec:MomentumBasis}) and explain how to implement this basis within the MeshGraphNets architecture (Sec. \ref{sec:MomentumGNN}). 

\subsection{Momentum Decomposition}
\label{sec:MomentumStep}
If a mechanical system is subject to \textit{external}, non-conservative forces due to, e.g., gravity or contact, the total momentum of the system will not be conserved. Nevertheless, the \textit{internal} forces arising, e.g., from an elastic potential will not induce any change in momentum. Using this observation, we decompose the integration of the system's equations of motions into two sub-steps. We first integrate the system forward in time using only inertia and non-conservative external forces. We then correct this \textit{momentum step} by adding momentum-conserving impulses predicted by a neural network.  

To implement this strategy, we recast the variational formulation of implicit Euler (\ref{eq:IELoss}) into a two-step update. Following \cite{Gast15,Brown18}, we define the momentum step as 
\begin{equation}
\bx^m = \bx^i +\Delta t \bv^i + \Delta t^2 \bM^{-1}\bff_\mathrm{ext} \ .     
\end{equation}
End-of-step position are then obtained as the minimizer of the modified implicit Euler potential,
\begin{equation}
\label{eq:modifiedIELoss}
    \min_\bx \frac{1}{2\Delta t^2}||\bx-\bx^m||^2_\bM + E_\mathrm{int}(\bx) \ .
\end{equation}
While this two-step version is equivalent to Eq. (\ref{eq:IELoss}), it provides the basis for a clean separation between {momentum-changing} accelerations due to external forces and {mo\-men\-tum-conserving} accelerations due to internal forces.
Specifically, we train our GNN to predict momentum-conserving accelerations $\ba_\mathrm{mc}$ such that the corresponding updated positions, 
\begin{equation}
    \bx^{i+1} = \bx^m + \Delta t^2\ba_\mathrm{mc} \ ,
\end{equation}
minimize the modified potential (\ref{eq:modifiedIELoss}) accumulated over all samples in the training set. While the end-of-step positions $\bx^{i+1}$ are generally not a true (i.e., unconstrained) minimizer of (\ref{eq:modifiedIELoss}), they constitute the momentum-conserving (i.e., constrained) solution that is closest to the implicit Euler step. With this basis laid out, we next describe how to build momentum-conservation into our GNN architecture.

\subsection{A Basis for Momentum-Conserving Impulses}
\label{sec:MomentumBasis}

To restrict our GNN to predict momentum-conserving accelerations by construction, we observe that the net change in linear momentum $\bp$ due to a given set of vertex accelerations $\ba_i$ is
\begin{equation}
    \frac{d\bp}{dt}= \sum_i \frac{d\bp_i}{dt} = \sum_i m_i\frac{d\bv_i}{dt} = \sum_i m_i\ba_i \ .  
\end{equation}
The last term in the above expression suggests that one way of achieving momentum conservation is to impose constraints requiring the mass-weighted sum of accelerations to vanish. Since enforcing hard constraints during training is difficult and can slow down learning significantly, we therefore seek to find a basis for the space of admissible impulses that guarantees momentum conservation by construction.

\paragraph{Momentum-Conserving Basis}

If we design impulses to only affect quantities invariant to translations and rotations, then such impulses cannot change linear or angular momentum. Let $q$ be an intrinsic quantity with Taylor expansion
\begin{equation}
    q(\bx+\Delta \bx)=q(\bx)+\nabla q(\bx)\cdot\Delta \bx+...
\end{equation}
Invariance implies that the first-order term vanishes for any $\Delta \bx$ corresponding to rigid transformation, i.e., $\nabla q(\bx)\cdot\Delta \bx=0$. The gradient of each of the six momentum components (three translational, three rotational), computed with respect to vertex positions, corresponds exactly to a rigid transformation direction (along xyz and around xyz). Therefore, $\nabla q$ is orthogonal to each momentum gradient, and impulses along $\nabla q$ preserve both linear and angular momentum.

Gradients of invariant quantities thus provide a basis for momentum-conserving impulses, but what quantities should be used? As a natural choice in our context, finite element discretizations and discrete differential geometry operators offer a large range of deformation measures that are invariant to rigid rotations and translations, such as areas and volumes. As the most basic option, we build our basis on edge lengths. To this end, we endow each edge with a single scalar $w_{ij}$ that describes the magnitude of an impulse along the gradient of edge length,
\begin{equation}
 \frac{\partial l(\bx_i, \bx_j)}{\partial \bx_i} = \frac{\bx_i-\bx_j}{|\bx_i-\bx_j|} \quad \text{and} \quad 
  \frac{\partial l(\bx_i, \bx_j)}{\partial \bx_j} = - \frac{\bx_i-\bx_j}{|\bx_i-\bx_j|} \ ,
\end{equation}
which, for each vertex, is simply the normalized edge vector. Edge impulses are then translated into pairs of per-vertex impulses as
\begin{equation}
    \Delta \bp_i = w_{ij}\frac{\partial l(\bx_i, \bx_j)}{\partial \bx_i} \quad \text{and} \quad \Delta \bp_j = - \Delta \bp_i \ .
\end{equation}
We collect corresponding basis vectors for each edge in a matrix $\bB$ and express mo\-men\-tum-conserving impulses as $\Delta \bp=\bB\bw$, where $\bw\in\mathbb{R}^{|E|}$ is a vector of per-edge impulse magnitudes. It is clear by construction that all impulse generated in this way are momentum-conserving. In the supplemental material, we show that this basis is likewise complete in the sense that any momentum-preserving impulse can be generated in this way.

The above formulation directly extends to the 3D case of volumetric objects represented using tetrahedral meshes. However, the case of triangle meshes embedded in 3D space---corresponding to thin shell materials such as cloth and paper---requires special attention.

\paragraph{Extensions to Thin Shells}
While 2D elastic sheets resist local changes in length, 3D thin shell materials must also respond to local changes in curvature, i.e., bending deformations. Our per-edge basis can generate any momentum-conserving in-plane responses, but it cannot create out-of-plane motion and must therefore be extended.
While different options are available in the literature, we follow \citet{Grinspun2003Discrete} and
\setlength{\columnsep}{8pt}%
\setlength{\intextsep}{8pt}%
\begin{wrapfigure}{r}{0.25\textwidth} 
    \centering
    \vspace{-0.3cm}
    \includegraphics[width=0.35\columnwidth]{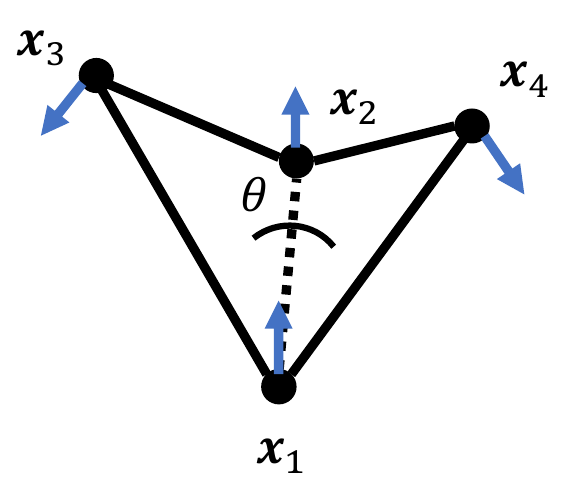}
    \caption{Dihedral angle $\theta$ defined by two edge-adjacent triangles. 
    %between two triangles with vertices ${\bf x}_1, {\bf x}_2, {\bf x}_3, {\bf x}_4$. 
    Per-vertex gradients $\frac{\partial \theta}{\partial {\bf x}_{i}}$ are shown in blue.}
    \label{fig:dihedral_angle}
    \vspace{-0.5cm}
\end{wrapfigure}
use the dihedral angle formed by two edge-adjacent triangles. As indicated in the inset figure, the dihedral angle is determined by the four vertex positions of its two adjacent triangles. To obtain momentum-conserving impulses that only change the dihedral angle (to first order), we endow each edge with an additional impulse magnitude variable $w^b_{ij}$ and define the four per-vertex impulses $\Delta \bp_i=w^b_{ij}\frac{\partial \theta}{\partial \bx_i}$. Closed form expressions for the corresponding derivatives can be found in \cite{WARDETZKY2007499}. Since the dihedral angle is invariant to rigid transformations, it follows that these impulses conserve linear and angular momentum.

\subsection{MomentumGNN} 
\label{sec:MomentumGNN}     
Implementing our momentum-conserving impulses requires only slight modifications to MeshGraphNets that replace per-vertex decoders with per-edge decoders.

\paragraph{Per-Edge Decoders} Specifically, we define an edge decoder MLP $f^\mathrm{stretch}$ that takes as input max-pooled edge features and predicts an impulse magnitude, \begin{equation}
    w^\mathrm{stretch}_{ij} = f^\mathrm{stretch}(\max({\bf s}_{ij}, {\bf s}_{ji})) \ , 
 \end{equation}
from which we then compute per-vertex impulses as
\begin{equation}
    \Delta{\bf p}_{ij}^\mathrm{stretch} =w_{ij}^\mathrm{stretch} \frac{\partial l_{ij}}{\partial \bx_i} \ .
\end{equation}
In an analogous way, we define an edge decoder MLP $f^\mathrm{bend}$ that predicts bending impulse magnitudes, \begin{equation}
    w_{ij}^\mathrm{bend}= f^\mathrm{bend}(\max({\bf e}_{ij}, {\bf e}_{ji})) \ ,
\end{equation}
from which we compute per-vertex impulses, 
\begin{equation}
    \Delta{\bf p}_{ij}^\mathrm{bend}=w_{ij}^\mathrm{bend} \frac{\partial \theta}{\partial {\bf x}_{i}} \ ,
\end{equation}
along the gradient of the dihedral angle $\theta_{ij}$ for edge $\be_{ij}$.
For each node, we sum the per-vertex contributions of all incident edges and transform the resulting impulses into nodal accelerations as  
\begin{align}
    {\bf a}_i &= \frac{1}{m_i \Delta t}\sum_{j} 
    \left(
    \Delta\bp_{ij}^\mathrm{stretch} + \Delta \bp _{ij}^\mathrm{bend} 
    \right ) \ .
    \label{eqn:acc_aggr}
\end{align}

\begin{wrapfigure}{r}{0.2\textwidth} 
    \centering
    \includegraphics[width=0.35\columnwidth]{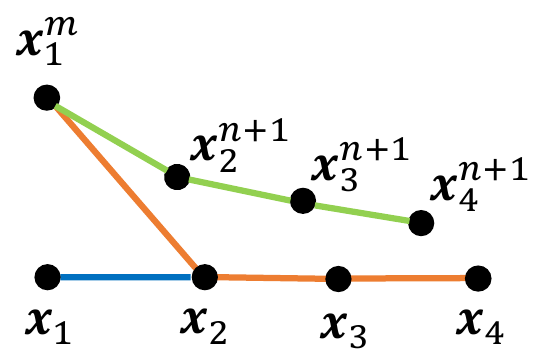}
    \caption{A 2D mass-spring chain. 
    }
    \label{fig:issue_demo}
    \vspace{-0.2cm}
\end{wrapfigure}

\paragraph{Position Updates}
Having replaced the per-vertex decoders with our momentum-conserving per-edge decoders, we can train the resulting GNN in the usual way. While the position updates predicted by this network conserve momentum, the space of possible updates is limited by the fact that all impulse directions are computed using the geometry of the momentum step $\bx^m$.  
To illustrate this limitation, we consider the example of a 2D mass-spring chain as shown in Fig. \ref{fig:issue_demo}. An external force acting on vertex $\bx_1$ leads to the momentum step predicting an updated position $\bx^m_1$ whereas the positions for the other nodes remain unchanged. When computing per-edge impulses based on this geometry (\textit{orange}), neither $\bx_3$ nor $\bx_4$ can be displaced vertically---but such vertical displacement is required to reach the implicit Euler solution indicated in green in Fig. \ref{fig:issue_demo}. 

\setlength{\columnsep}{8pt}%
\setlength{\intextsep}{8pt}%

While smaller step sizes will alleviate this problem, its root cause is the fact that, while each message passing layer adjusts all impulse magnitudes, the direction of these impulses is kept constant across all steps. A natural way of increasing the space of momentum-conserving impulses is therefore to update the geometry with the current impulses after each message passing step. Since each such per-layer update conserves momentum, any sequence of per-layer updates will likewise conserve momentum.
We implement this layered position update strategy as shown in Figure \ref{fig:overview}. The resulting \textit{MomentumGNN} architecture decodes and updates positions for each layer (indicated in orange). The positions output by the last layer ${\bf x}^{(L)}$ are used as the final prediction ${\bf x}^{i+1}$. 
In order to feed these position updates back into the message passing process, we explicitly recompute dihedral angles $\theta^l_{ij}$ and edge strains $\varepsilon^l_{ij}$ from the intermediate positions ${\bf x}^{(l)}$ and modify the edge MLP from Eq. (\ref{eqn:message}) as 
\begin{align}
    {\bf m}^{l+1}_{ij}\leftarrow f^\mathrm{edge}({\bf s}_{ij},{\bf r}_i,{\bf r}_j, \theta^l_{ij}, \varepsilon^l_{ij}) \ .
\end{align}

\paragraph{Velocity Projection}
Our network predicts momentum-conserving position updates, but it does not directly return updated velocities. While one could use the finite difference approximation of (\ref{eq:velFD}) to compute velocities $\bv^\mathrm{FD}$, those velocities would generally not conserve angular momentum when evaluated on the updated positions.
We therefore compute velocities by finding the smallest correction to $\bv_i^\mathrm{FD}$ that preserves momentum. We cast this problem as a convex quadratic program,
\begin{align}
    \min_{\Delta \bv} \ \frac{1}{2}\|\Delta \bv\|^2 
    \quad \text{s.t.}\quad & \sum_i m_i (\bv_i^\mathrm{FD}+\Delta \bv_i)={\bf p}^t\\
    &\sum_i m_i {\bf x}_i\times(\bv_i^\mathrm{FD}+\Delta \bv_i)={\bf L}^t \ .
\end{align}
The Lagrangian of this optimization problem is
\begin{align}
    L&=\frac{1}{2}\|\Delta \bv\|^2+\boldsymbol{\lambda}^T {\bf C}(\bv^\mathrm{FD} + \Delta\bv) \ ,
\end{align}
where $\bC$ summarizes all 6 constraints. The first-order optimality conditions require the gradient of the Lagrangian with respect to $\bv$ and $\boldsymbol{\lambda}$ to vanish, which leads to the linear system
\begin{align}
    \begin{bmatrix}
    \bf M & \nabla \bC^T \\
    \nabla \bC & \bf 0
    \end{bmatrix}\begin{bmatrix}
    \Delta \bv\\
    \boldsymbol{\lambda} 
    \end{bmatrix}=\begin{bmatrix}
    \bf 0\\ 
    -\bC(\bv^\mathrm{FD})
    \end{bmatrix} \ .
\end{align}
For efficiency, we first compute Lagrange multipliers $\boldsymbol{\lambda}$ by solving the $6\times 6$ linear system
\begin{equation}
    \nabla\bC{\bf M}^{-1}\nabla\bC^T \boldsymbol{\lambda} =\bC(\bv^\mathrm{FD}) \ ,
\end{equation}
and then compute momentum-corrected velocities as
\begin{align}
    \bv^{t+1}&=\bv^\mathrm{FD}-{\bf M}^{-1}\nabla\bC^T\boldsymbol{\lambda} \ .
\end{align}

\section{Results}
\subsection{Training scheme}
We implement our method in PyTorch \cite{paszke2019pytorch} and perform training and testing on a single RTX 4090 GPU. Each GNN layer is implemented by a 2-layer MLP with 128 latent dimensions, layer normalization, and Gaussian Error Linear Units (GELU) as activation functions. We use discrete shells and constant strain triangles with a Saint Venant-Kirchhoff material for cloth and a Neo-Hookean material for volumetric solids. We summarize our experiment setup in Table~\ref{tab:exp}.

We generate training data procedurally using a Python script that collects sequences of simulation snapshots of random cuboids and sheets recovering from initial deformations. We randomly sample tetrahedral cuboids and sheets of different sizes using the Gmsh package. We then randomly deform cuboids by affine transformations and sheets along Bezier curves. Cuboids and sheets are assigned random initial velocities.

\begin{table}[t]
\caption{Experimental setup.}

\vspace{-0.3cm}
\label{tab:exp}
\begin{minipage}{\linewidth}
\begin{center}
\resizebox{\linewidth}{!}{
\begin{tabular}{lcccccc}
  \toprule
  Experiment & \# of GNN layers & \# of vertices & \# of cells & Timestep \\
  \midrule
    Falling cloth (torus) & 10 & 2119 & 4068 & 11  \\
    Swinging and falling cloth  & 20 & 514 & 946 & 21 \\
  Shooting baskets  & 20 & 993 & 4078 & 17 \\
  Bouncing tennis ball  & 20 &  396 & 1480 & 12 \\
  Armadillo & 20 & 10281 & 55091 & 82 \\
  \bottomrule
\end{tabular}
}
\end{center}
\end{minipage}

\vspace{-0.3cm}
\end{table}

Following recent work  \cite{datafree,snug,grigorev2022hood,ncs}, we train our GNN in a self-supervised way using a physics-based loss function that sums up the per-step potential from (\ref{eq:modifiedIELoss}).
We use the Adam optimizer with a learning rate $1 \times 10^{-5}$ and add two types of noise to the training data. The first training noise is applied to all vertices and sampled from a normal distribution $\mathcal{N}(0,\sigma)$ ($\sigma=0.001$ for cloth and $\sigma=6 \times 10^{-4}$ for solid). The second training noise is applied to a random local neighborhood within a radius of $1/4$ (cloth) or $1/2$ (solid) of the longest edge of the sheet or cuboid. We sample the same direction for all vertices and independently sample magnitudes from $\mathcal{U}(0,0.05)$. 
For a fair comparison, when training self-supervised MeshGraphNets baselines, we additionally sample random rotations for both rest and deformed configurations, as they are not equivariant to rotations.

We collect additional data for the pinned cloth. We randomly sample gravity-like external forces of uniform direction and a magnitude from a uniform distribution $\mathcal{U}(0,20)$. We collect the snapshots by releasing the pinned cloth from its rest state. We finetune pre-trained cloth checkpoints by jointly training on both pinned and unpinned data.

\begin{figure}[ht]
% \vspace{0.1cm}
\begin{minipage}{\columnwidth}
    \centering
    \includegraphics[width=\columnwidth]{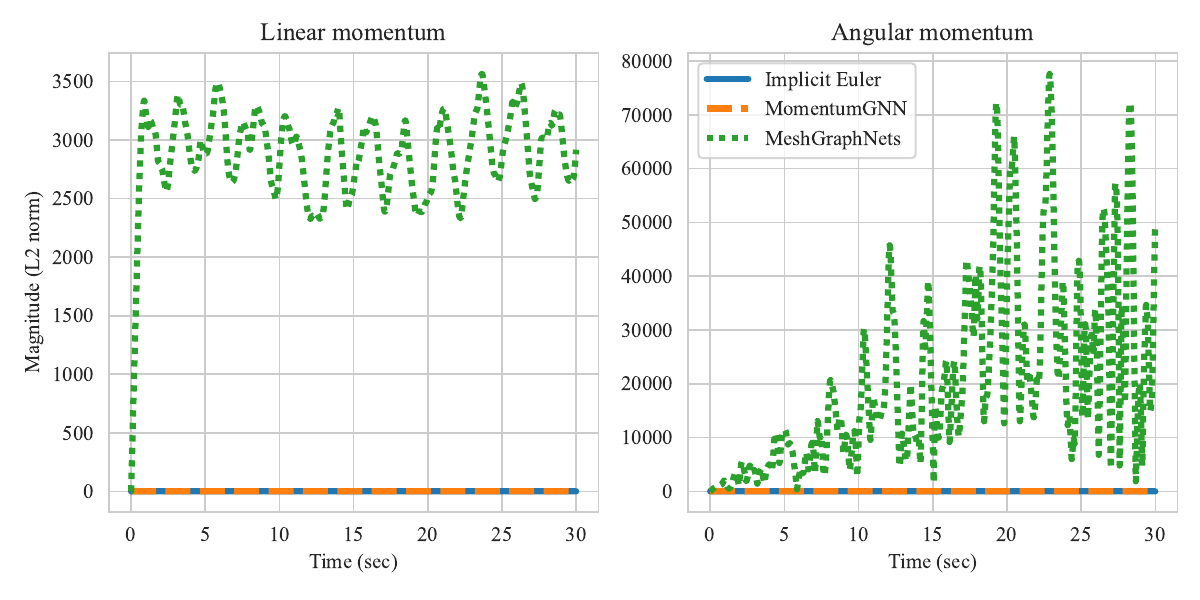}
    \vspace{-0.9cm}
    \caption{MomentumGNN conserves linear and angular momentum of the Armadillo, while MeshGraphNets quickly leads to an explosion.}
    \label{fig:plot_armadillo_momentum}
\end{minipage}
\vspace{-0.3cm}
\end{figure}
\begin{figure}[ht]
\vspace{-0.1cm}
\begin{minipage}{\columnwidth}
    \centering
    \includegraphics[width=\columnwidth]{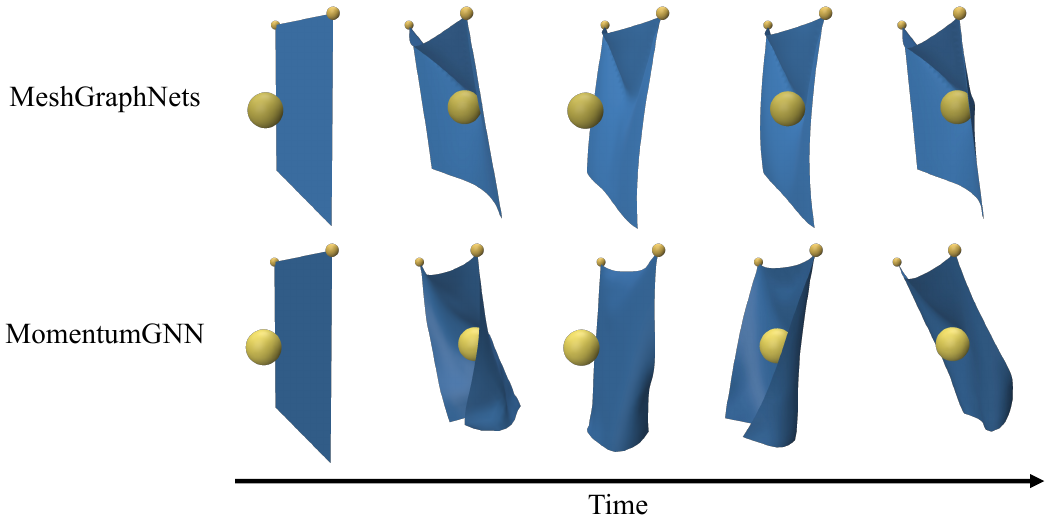}
    \vspace{-0.6cm}
    \caption{We simulate a hanging cloth interacting with a moving ball.}
    \label{fig:cloth_sphere}
\end{minipage}
% \vspace{-0.3cm}
\end{figure}
\begin{figure}[ht]
\vspace{-0.5cm}
\begin{minipage}{\columnwidth}
    \centering
    \includegraphics[width=0.78\columnwidth]{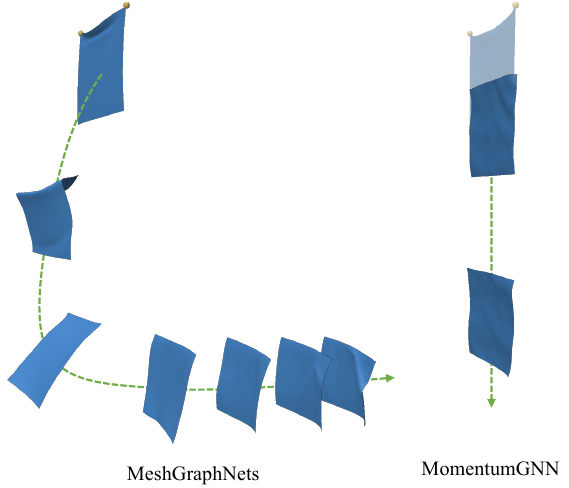}
    \vspace{-0.3cm}
    \caption{We release the pinned-vertex constraints for the hanging cloth and simulate its falling motion.}
    \label{fig:falling_cloth}
\end{minipage}
\vspace{-0.5cm}
\end{figure}
\subsection{Swinging and Falling Cloth}
We begin our evaluation by comparing MomentumGNN (our method) with MeshGraphNets on cloth simulation. We train MeshGraphNets using L2 supervision, following their distributed codebase and dataset. As shown in Figure \ref{fig:cloth_sphere}, both models achieve comparable quality in simulating a swinging cloth. However, MomentumGNN produces more dynamic and realistic motion, attributed to its explicit conservation of momentum.
The limitations of MeshGraphNets become apparent under slight scene modifications. In Figure \ref{fig:falling_cloth}, we simulate a falling cloth by removing the pins. MeshGraphNets does not generalize to this new scenario. Rather than falling vertically under gravity, the cloth drifts horizontally and exhibits unnatural momentum. This discrepancy is more evident in the supplementary video.
In contrast, MomentumGNN accurately captures the expected physical behavior. It accurately predicts the evolution of momentum---in particular zero linear momentum in lateral directions---enabling the cloth to fall straight downward while naturally forming wrinkles during descent.

\begin{figure}[ht]
% \vspace{0.1cm}
\begin{minipage}{\columnwidth}
    \centering
    \includegraphics[width=\columnwidth]{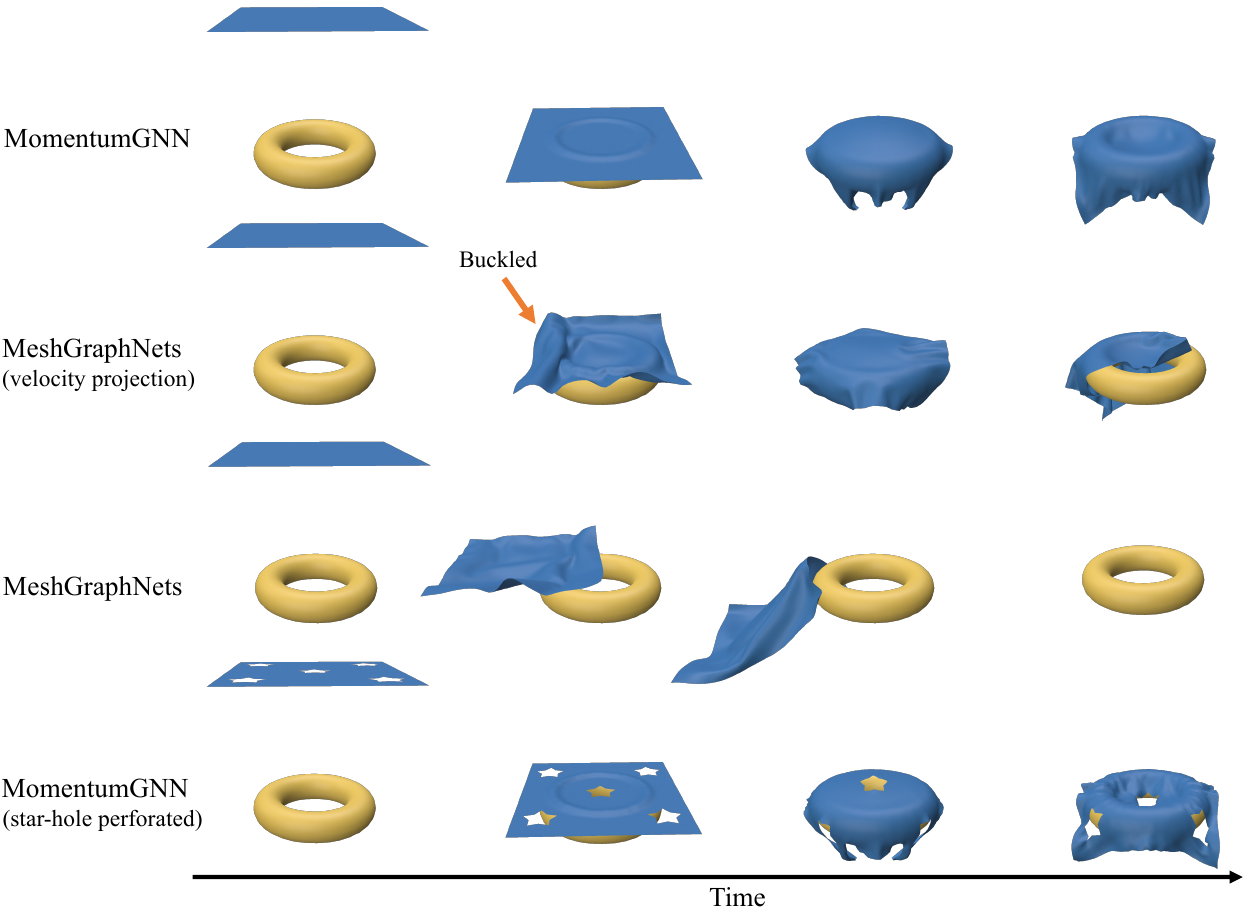}
    \vspace{-0.7cm}
    \caption{We drop a cloth onto a torus and simulate the cloth dynamics. We also showcase an extension to star holes (genus 5).}
    \label{fig:cloth_torus}
\end{minipage}
\vspace{-0.3cm}
\end{figure}
\begin{figure*}[ht]
\vspace{-0.7cm}
\begin{minipage}{\linewidth}
    \centering
    \includegraphics[width=0.9\linewidth]{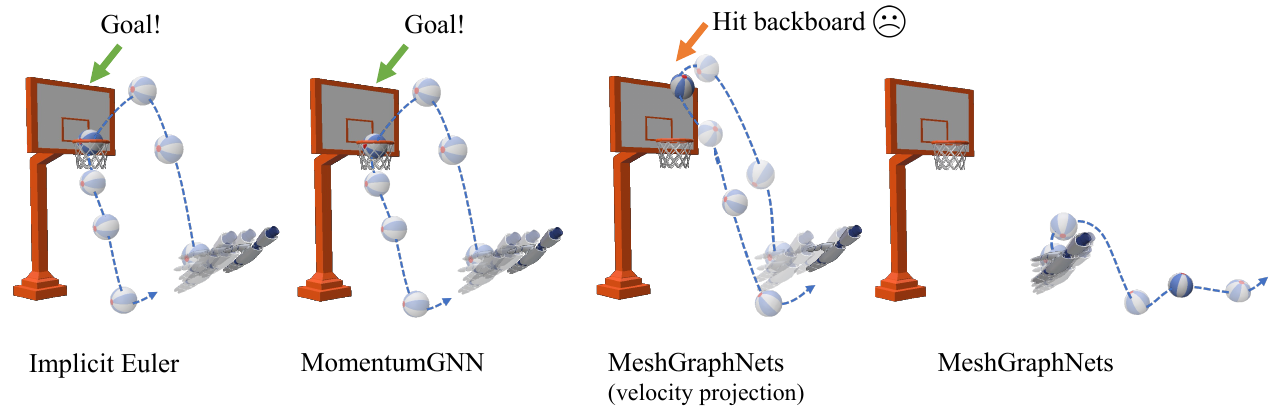}
    \vspace{-0.3cm}
    \caption{We simulate a basket-shooting scene, where a robotic hand throws a beach ball to the basket. Both MomentumGNN and Implicit Euler successfully complete the throw, whereas MeshGraphNets fails. The improvement with the velocity projection directs the ball toward the basket but does not score.}
    \label{fig:basket}
\end{minipage}
\vspace{-0.6cm}
\end{figure*}
\subsection{Falling Cloth on Torus}
To further validate this observation, we retrain MeshGraphNets on the same dataset as our model using self-supervision. When dropping a cloth onto a torus (Figure \ref{fig:cloth_torus}), MeshGraphNets again exhibits drift and does not drape on the torus.
We further introduce an enhanced baseline by replacing MeshGraphNets’ finite-difference velocity with our velocity projection formulation. 
This modification mitigates drift but introduces buckling artifacts, indicating that velocity projection alone is insufficient, and highlighting the importance of coherent momentum modeling throughout the architecture.
In contrast, MomentumGNN produces stable, physically-plausible trajectories, accurately capturing both free-fall dynamics and visually plausible folding after contact, and generalizes to perforated star shapes.

\begin{figure}[ht]
% \vspace{0.1cm}
\begin{minipage}{\linewidth}
    \centering
    \includegraphics[width=0.95\linewidth]{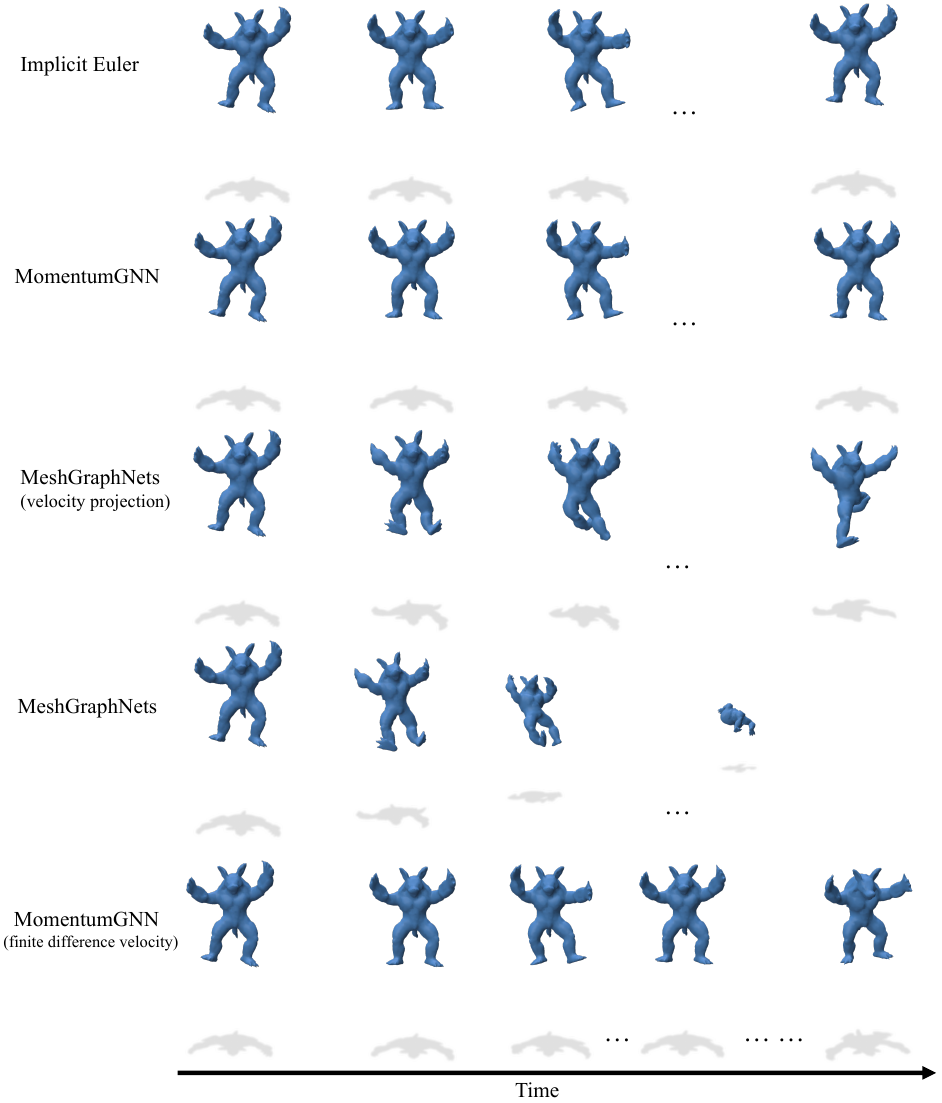}
    \vspace{-0.3cm}
    \caption{We simulate an Armadillo released from a nonlinear deformed pose with zero initial velocity and no external forces.}
    \label{fig:armadillo}
\end{minipage}
\vspace{-0.5cm}
\end{figure}
\subsection{Shooting Baskets}
We evaluate MomentumGNN on elastic solid simulation using a basket-shooting scenario, where a robotic hand throws a ball toward a hoop (Figure \ref{fig:basket}). As shown in the figure and the supplementary video, MomentumGNN faithfully captures the ball’s natural trajectory, successfully completing the shot. In contrast, MeshGraphNets fails to reproduce the correct motion, with the ball veering off course due to spurious momentum injection. Augmenting MeshGraphNets with velocity projection partially corrects the trajectory, directing the ball toward the basket, but the shot ultimately strikes the backboard and misses.

\subsection{Armadillo}
Despite being trained exclusively on simple cuboids, MomentumGNN generalizes effectively to complex shapes, such as Armadillo (see Figure \ref{fig:armadillo}). The Armadillo is initially deformed into a nonlinear pose with zero initial velocity and then allowed to evolve freely without external forces. MomentumGNN accurately captures the natural swinging motion of the arms and legs. In contrast, MeshGraphNets exhibits drift, rotating the Armadillo arbitrarily out of the scene, violating momentum conservation. Although augmenting MeshGraphNets with velocity projection helps maintain the Armadillo within the scene, it still is incapable of producing physically correct dynamics. These differences are best seen in the supplementary video.

We also provide a quantitative evaluation by computing the linear and angular momentum of the Armadillo, plotted in Figure \ref{fig:plot_armadillo_momentum}. Since the Armadillo begins at rest with no external forces applied, both momenta should remain zero throughout the simulation. As illustrated in Figure \ref{fig:plot_armadillo_momentum}, MeshGraphNets exhibits rapid growth in momentum, diverging from the expected behavior. In contrast, MomentumGNN faithfully conserves momentum with only a modest computational overhead compared to MeshGraphNets (12 fps vs. 14 fps), while offering a substantial performance improvement over implicit Euler (2 fps).

Finally, we conduct a brief ablation study by replacing our momentum-consistent velocity with the finite difference velocity during simulation. As shown in Figure \ref{fig:armadillo}, MomentumGNN still effectively captures the swinging motion of the arms and legs, but, as expected, produces gradual deviations in orientation.
\section{Conclusion}

We presented a graph-based neural architecture for simulating deformable objects that explicitly conserves linear and angular momentum. Our method predicts momentum-conserving accelerations by modeling stretching and bending impulses via per-edge decoders. To enhance the representational capacity of the network, we introduced a layer-by-layer update scheme that allows richer hierarchical processing of deformation dynamics. Trained on canonical geometries such as sheets and cuboids, our network generalizes effectively to a wide array of challenging scenarios involving cloth and soft solids. Experimental results showcased our model's ability to capture rich deformation patterns while preserving the momentum of the system.

{\it Limitations and future work.} We did not explore hierarchical message passing, which we consider a complementary and promising direction. Moreover, although our networks can, in principle, support variations in material parameters, we have not conducted any experiments in this direction; integrating learned constitutive laws \cite{ma2023learning} from observational data represents an exciting avenue to discover novel material behaviors within a unified framework.

\newpage
{
    \small
    \bibliographystyle{ieeenat_fullname}
    \bibliography{reference}
}
\clearpage
\setcounter{page}{1}
\appendix
\setcounter{table}{0}
\setcounter{figure}{0}
\renewcommand{\thetable}{A\arabic{table}}
\renewcommand\thefigure{A\arabic{figure}}
\renewcommand{\theHtable}{A.Tab.\arabic{table}}%<---!!!!---
\renewcommand{\theHfigure}{A.Abb.\arabic{figure}}%<---!!!!---
\renewcommand\theequation{A\arabic{equation}}
\renewcommand{\theHequation}{A.Abb.\arabic{equation}}%<---!!!!---
% \maketitlesupplementary
\newpage
\twocolumn[
        \centering
        \Large
        \textbf{\thetitle}\\
        \vspace{0.5em} Appendix \\
        \vspace{1.0em}
       ]

\section{Completeness of Per-Edge Momentum Basis}
We show that per-edge momentum basis is complete in the sense that any momentum-preserving impulse can be generated in this way.

To this end, we assume that there exist momentum-conserving impulses impulses $\bq$ that cannot be generated using the basis $\bB$, i.e., 
\begin{equation}
    \sum_i \bq_i = \mathbf{0} \ , \quad
    \sum_i (\bq_i\times\bc) = \mathbf{0} \ , \ \ 
    \text{and} \ \  \bB\bB^T\bq-\bq=\bb \neq \mathbf{0} \ ,
\end{equation}
where $\bc$ is an arbitrary reference point.
Without loss of generality, we consider a two-dimensional example and show by induction that $\bB$ always has rank $2|V|-3$ with the three-dimensional null-space of $\bB^T$ corresponding to rigid transformations. To see this, consider a single triangle with six nodal degrees of freedom. The three gradients of edge length are linearly independent vectors and thus form a three-dimensional subspace of $\mathbb{R}^6$. Since translations and (first-order) rotations are orthogonal to these gradients, the set of vectors $\bv$ for which $\bB^T\bv=\mathbf{0}$ is a three-dimensional null-space corresponding to (first-order) rigid transformations. Now consider adding a vertex to an existing triangle mesh. This vertex adds two degrees of freedom, but we must also add at least two additional edges to maintain a proper triangulation. The two additional edges yield two linearly independent basis vectors because the corresponding edges are not collinear. Since the new basis vectors involve the new vertex, they are also linearly independent of all pre-existing basis vectors. Consequently, the output dimension of $\bB$ increases by 2, but so does its rank, implying that the dimensionality and structure of the null-space remains unchanged. 

Using this observation, it is clear that either $\bB^T\bb\neq\mathbf{0}$ or that $\bb$ corresponds to a first-order rigid transformation and therefore changes momentum. We therefore conclude that all momentum-conserving impulses can be expressed with the per-edge basis $\bB$.

\section{Discussion on method}
\paragraph{Impulse Basis from Intrinsic Properties instead of Energy}
We prefer to think in terms of intrinsic properties such as edge lengths and dihedral angles because their gradients are always well-defined and nonzero. Elastic energies due to stretching and bending build on these intrinsic properties but are at least quadratic functions of them, such that their gradients vanish at rest (when there is no deformation), resulting in degenerate impulse directions.

\paragraph{Momentum Conservation and Dissipation}
Our method predicts nodal updates for a momentum step that preserve total linear and angular momentum. It is important to note that momentum conservation does not imply the conservation of energy. To see this, consider a discretized elastic bar that is stretched along its axis and then released. As the bar oscillates, its total linear momentum remains zero--- regardless of the amplitude oscillation. It is evident that scaling all nodal velocities by a given constant does not change the total momentum. However, the kinetic energy of the system can be changed arbitrarily in this way.
Since momentum-conserving impulses can dissipate energy, our method learns the numerical damping inherent to implicit Euler even when using a purely elastic material.

\paragraph{Implicit Euler as Basis}
It is well established that implicit Euler conserves neither energy nor momentum. While this seemingly disqualifies implicit Euler as a basis for our momentum-preserving GNN, its stability properties nevertheless make it a very attractive choice. For example, while symplectic Euler conserves momentum by construction, it has poor stability properties. Similarly, the implicit midpoint scheme preserves momentum but does not enjoy the same stability properties as implicit Euler. By building our loss function on implicit Euler, our GNN learns to predict momentum-preserving corrections that lead to stable behavior even for very long roll-outs. 

\section{Additional results}
\subsection{Bouncing Tennis Ball}
\begin{figure*}[ht]
% \vspace{0.1cm}
\begin{minipage}{\linewidth}
    \centering
    \includegraphics[width=0.95\linewidth]{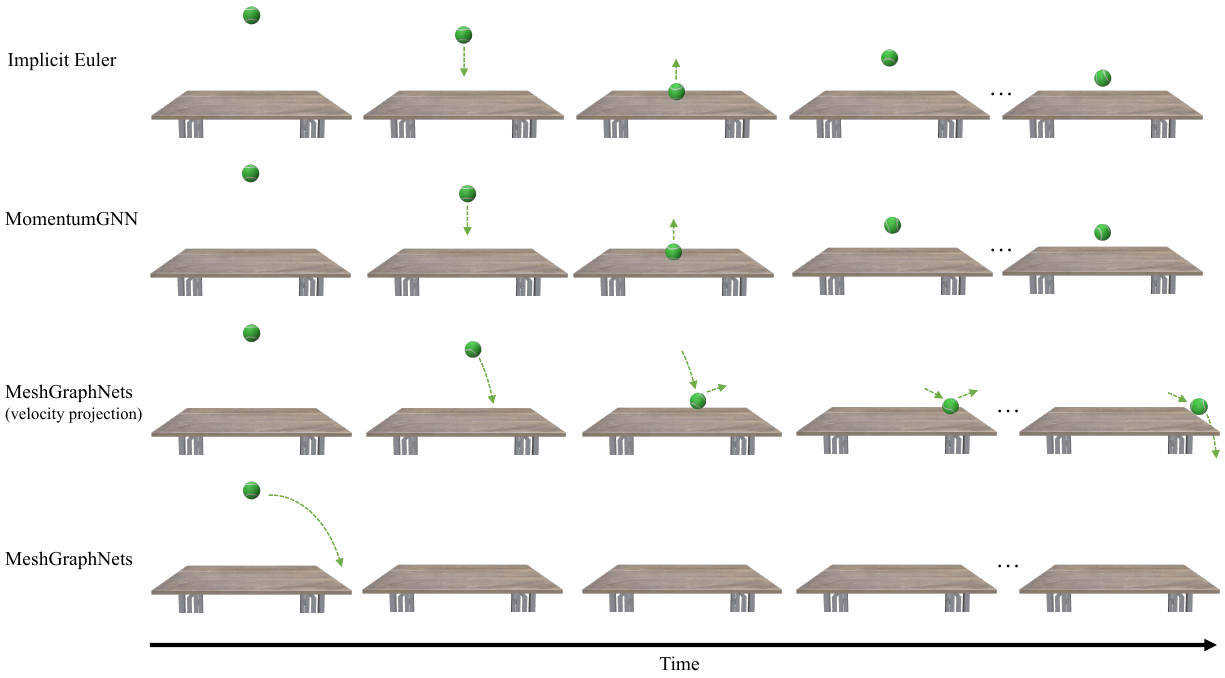}
    \caption{We simulate a bouncing tennis ball on a table. While MomentumGNN and Implicit Euler produce natural bouncing motions, both MeshGraphNets drifts the ball off the table.}
    \label{fig:tennis}
\end{minipage}
% \vspace{-0.3cm}
\end{figure*}

We extend our evaluation by simulating a tennis ball dropped onto a table. As shown in Figure \ref{fig:tennis}, our approach closely matches the behavior of implicit Euler, generating realistic bouncing behavior. Additionally, it accurately captures the damping effects observed in the reference simulation, resulting in a gradual reduction of rebound amplitude over time. By contrast, MeshGraphNets is not able to generate physically plausible motion for the ball. Incorporating velocity projection improves the trajectory, but the ball ultimately bounces off the table due to spurious momentum changes.

\end{document}